# Directional Deep Embedding and Appearance Learning for Fast Video Object Segmentation

Yingjie Yin, De Xu, *Senior Member, IEEE*, Xingang Wang and Lei Zhang, *Fellow, IEEE*

*Abstract*—Most recent semi-supervised video object segmentation (VOS) methods rely on fine-tuning deep convolutional neural networks online using the given mask of the first frame or predicted masks of subsequent frames. However, the online fine-tuning process is usually time-consuming, limiting the practical use of such methods. We propose a directional deep embedding and appearance learning (DDEAL) method, which is free of the online fine-tuning process, for fast VOS. First, a global directional matching module, which can be efficiently implemented by parallel convolutional operations, is proposed to learn a semantic pixel-wise embedding as an internal guidance. Second, an effective directional appearance model based statistics is proposed to represent the target and background on a spherical embedding space for VOS. Equipped with the global directional matching module and the directional appearance model learning module, DDEAL learns static cues from the labeled first frame and dynamically updates cues of the subsequent frames for object segmentation. Our method exhibits state-of-the-art VOS performance without using online fine-tuning. Specifically, it achieves a $\mathcal{J}$ & $\mathcal{F}$ mean score of 74.8% on DAVIS 2017 dataset and an overall score $\mathcal{G}$ of 71.3% on the large-scale YouTube-VOS dataset, while retaining a speed of 25 fps with a single NVIDIA TITAN Xp GPU. Furthermore, our faster version runs 31 fps with only a little accuracy loss. Our code and trained networks are available at https://github.com/YingjieYin/Directional-Deep-Embedding-and-Appearance-Learning-for-Fast-Video-Object-Segmentation.

## I. INTRODUCTION

Video object segmentation (VOS) is an important problem in the field of computer vision and it is widely used in applications such as video editing, robot visual perception and human-computer interaction, etc. In this work, we focus on the task of semi-supervised VOS which tracks and segments one or multiple objects whose ground-truth segmentation masks are given in the first frames of the videos. Owe to the great successes of Deep Convolutional Neural Networks (DCNNs) in image segmentation [1][2][3], visual tracking [4][5][6] and object detection [7][8][9], most of the recent high-accuracy semi-supervised VOS methods rely on fine-tuning a DCNN online to learn the appearances of the target objects. Unfortunately, online fine-tuning-dependent VOS methods usually have a slow speed because the fine-tuning process costs much extra time. This largely limits their usage in practical applications.

To tackle the aforementioned problem of online fine-tuning-dependent VOS methods, recent studies have focused on designing fine-tuning-free network architectures which can completely avoid online optimization. PML[10] and FEELVOS [11] used pixel-wise embedding learning in the Euclidean space to learn pixel discriminative features and internal guidance, respectively, for fine-tuning-free VOS. However, pixel-wise embedding learning in the Euclidean space needs a lot of calculations for similarity matching, which is time-consuming. The A-GAME [12] achieves fine-tuning-free VOS by learning the target appearance with high dimensional deep features in the Euclidean space. However, learning probabilistic generative model in the high dimensional Euclidean space faces the problem of "curse of dimensionality" [50], which attenuates the representation of target and background for effective segmentation.

In this work, we propose a directional deep embedding and appearance learning (DDEAL) method for fine-tuning-free fast VOS. Different from previous pixel-wise embedding learning methods in the Euclidean space, DDEAL employs directional deep features on a hyper sphere space to achieve pixel-wise embedding learning. The mixture of von Mises–Fisher (vMF) distribution [42] on a hyper sphere is used as the generative probabilistic model for the directional features of the foreground and background. Compared with the appearance model described by the mixture of Gaussians in the Euclidean space such as the one in A-GAME [12], our directional appearance model described by the mixture of vMF on a hyper sphere can more robustly represent the foreground and background's appearances and attenuate the influence of the "curse of dimensionality".

The main contributions of this paper are as follows.

(1) An online fine-tuning-free method, namely directional deep embedding and appearance learning (DDEAL), is developed for fast VOS. Directional static cues and dynamically updated directional cues are learned as a whole to achieve fast and accurate VOS.

(2) A global direction matching module is proposed to learn a semantic pixel-wise embedding with static cues from the la-

Y. Yin is with the Research Center of Precision Sensing and Control, Institute of Automation, Chinese Academy of Sciences, Beijing 100190, China, and also with the Department of Computing, The Hong Kong Polytechnic University, Hong Kong, China, and also with the School of Artificial Intelligence, University of Chinese Academy of Sciences, Beijing 100049, China. (e-mail: yingjie.yin@ia.ac.cn).

D. Xu and X. Wang are with the Research Center of Precision Sensing and Control, Institute of Automation, Chinese Academy of Sciences, Beijing 100190, China, and also with the School of Artificial Intelligence, University of Chinese Academy of Sciences, Beijing 100049, China. (de.xu@ia.ac.cn; xingang.wang@ia.ac.cn).

L. Zhang is with the Department of Computing, The Hong Kong Polytechnic University, Hong Kong. (e-mail: cslzhang@comp.polyu.edu.hk).

42beled first frame. This module can be efficiently implemented by convolution operations.

(3) An effective directional appearance model is proposed to represent the target and background on a spherical embedding space, which can effectively learn dynamic cues from the subsequent frames of the video for VOS.

To the best of our knowledge, this is the first work to learn directional cues for VOS. Extensive experiments on three benchmark datasets, including DAVIS 2016 [13], DAVIS 2017 [14] and the recent large-scale YouTube-VOS dataset [15], demonstrated that our proposed DDEAL achieves new state-of-the-art for multi-object video segmentation with faster speed than competing methods.

II. RELATED WORKS

A. *Semi-supervised video object segmentation*

Fine-tuning DCNNs online is adopted by many recent semi-supervised VOS methods to achieve higher accuracy [16][17][18][19][20][21][22][23]. These methods usually rely on training on synthetic and augmented data using the given mask of the first frame or predicted masks of previous frames at test time. OSVOS [16] fine-tunes a pre-trained DCNN on the annotated first-frame to segment the rest frames. OnAVOS [18] employs the segmentation results during the test time as new training examples to update the network online. LucidTracker [21] trains segmentation networks using in-domain synthetic training data generated by the provided annotations of each video's first frame to generalize across domains. Though achieving impressive segmentation accuracy, fine-tuning-dependent VOS methods cost much time in online learning during the testing stage, which largely deteriorates their practicability.

In order to address the problem of slow running speed of fine-tuning-dependent VOS methods, recent VOS studies have been concentrated on fine-tuning-free methods [10] [11] [12] [24] [25] [26] [27] [28] [29] [30], aiming at faster runtime and better usability. Cheng et al. proposed FAVOS [24], where a ROI segmentation network is developed to accurately output partial object segmentations. RGMP [25] predicts the target object' masks by reference-guided mask propagation. FEELVOS [11] uses the nearest neighbor matching results of pixel-wise embedding features as an internal guidance of the DCNN for VOS. Johnander et al. [12] proposed a network called A-GAME that learns a generative probabilistic appearance model of the target and background in the high dimensional Euclidean space for VOS. Compared with A-GAME, our proposed DDEAL performs embedding learning and appearance learning using directional features in the spherical embedding space, and it is more efficient and more robust.

B. *Deep embedding learning*

Deep embedding learning has been used in face recognition and verification [31][32], Image Segmentation [33], image retrieval and clustering [34][35] and image understanding [36]. Liu et. al. [32] proposed a deep hypersphere embedding approach for face recognition, which employs the angular softmax loss to learn discriminative face features with angular margin. A deep embedding learning method was proposed in [33] which can efficiently convert super-pixels into image segmentation. Chen et. al. [34] proposed an adaptive large margin N-pair loss to produce discriminative embedding under heterogeneous feature distribution for image retrieval and clustering. Li et al. [36] proposed a deep collaborative embedding model for multiple image understanding tasks. In our proposed DDEAL, pixel-wise embedding learning is proposed to learn static cues from the labeled first frame for VOS.

C. *Directional statistics-based learning*

The vMF distribution is a probability distribution for directional features, which are unit length vectors corresponding to points on a hyper sphere. The vMF distribution has been used for face verification [37], clustering [38], image classification and retrieval [39], machine translation [40] and document classification [41]. Hasnat et. al. [37] proposed a directional feature representation model based on the vMF mixture distribution for face verification. Gopal et al. [38] developed vMF based models for clustering high-dimensional data on a unit sphere as an alternative to the multinomial or Gaussian distribution based models. Kumar et al. [40] proposed probabilistic loss functions based on vMF distribution for sequence to sequence learning for language generation. In our proposed DDEAL, a directional appearance model based on vMF distribution is learned to provide dynamically updated cues for VOS.

III. METHODOLOGY

This work aims to develop a fine-tuning-free DCNN for VOS by using directional embedding and appearance learning. The directional embedding learning learns static cues from the labeled first frame, while the directional statistics based appearance model dynamically updates the cues from the following video frames. The complementary static and dynamic cues are integrated into a segmentation network which is trained in an end-to-end manner.

A. *Overview of the architecture*

Our DDEAL shares a similar structure to A-GAME [12], where there are a feature extractor module, a mask-propagation module [25], a fusion module and an up-sampling module. Different from A-GAME, there are two key directional feature based modules in DDEAL: the global directional matching module and the directional appearance module. The architecture of DDEAL is shown in Fig. 1. The features of the first frame and the input frame are extracted by a backbone network such as ResNet50 or ResNet101. The given mask of the first frame is resized to the same size as the extracted features. Then the extracted features and the resized mask are passed to the mask-propagation module, the global direction matching module and the directional appearance module. The outputs of these three modules are concatenated and input into the fusion module, using two convolutional layers to generate fused feature maps. A coarse predictor, which is implemented on the output of the fusion module by one convolutional layer, outputs a coarse segmentation mask. This coarse prediction is recycled by the mask-propagation layer in the next frame to update the appearance module. An up-sampling module and a predictor are also implemented on the output of the fusion module to output the final segmentation mask with the same size as the input frame. In addition, low-level information contained in the shallow features is fused in the up-sampling module by skip concatenation.

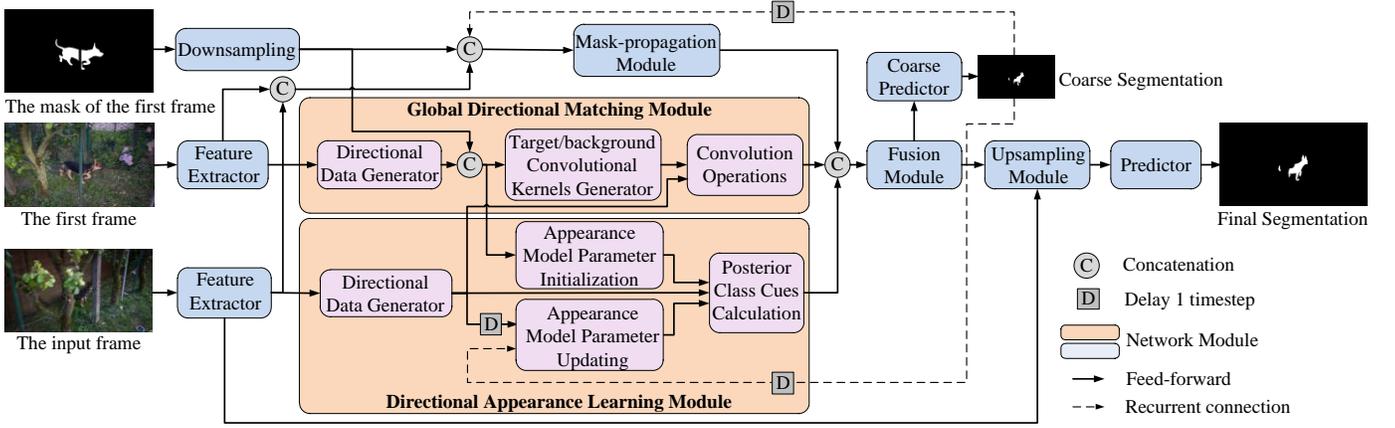

Fig. 1. Architecture of the proposed DDEAL approach for fast VOS. Two key directional feature based modules: global directional matching module and directional appearance learning module are designed. The former learns static cues from the labeled first frame and the latter learns dynamically updated cues from the following frames of the video.

### B. Embedding learning based on global direction matching

We use embedding learning with global directional matching to learn static cues from the labeled first frame. The process of global directional matching for the target object is shown in Fig. 2. Different from pixel-wise embedding learning based on Euclidean distance in previous methods, the cosine similarity is used in our global directional matching. For a given object or its background, we match the embedding directional features between the current frame and the first frame to produce a cosine similarity map. In the Euclidean space, the matching process needs to calculate the L2-norm of a lot of feature pairs, which is difficult to be accelerated directly through existing deep learning frameworks. We thus use cosine similarity metric of directional features in the spherical embedding space to perform the matching process, which can be efficiently implemented by convolution operations.

Denote by $F^0 \in \mathbb{R}^{1 \times C \times H \times W}$ the normalized feature maps extracted from the first frame, which consists of a set of directional feature vectors. A directional feature vector is with unit $L_2$-norm, corresponding to a point on a hyper sphere. Let $M^0 \in \mathbb{R}^{1 \times 1 \times H \times W}$ represent the resized mask of the object in the first frame. To generate the convolutional kernels $K^t \in \mathbb{R}^{HW \times C \times 1 \times 1}$ and $K^b \in \mathbb{R}^{HW \times C \times 1 \times 1}$, which include features of the object and background, respectively, we let

$$K^t_{hw,c,1,1} = M^0_{1,1,h,w} \cdot F^0_{1,c,h,w} \quad (1)$$

$$K^b_{hw,c,1,1} = (1 - M^0_{1,1,h,w}) \cdot F^0_{1,c,h,w} \quad (2)$$

where $K^t_{(hw),c,1,1}$ and $K^b_{(hw),c,1,1}$ are the values of $K^t$ and $K^b$, respectively, at position $(hw,c,1,1)$. The size of $K^t$ and $K^b$ is $HW \times C \times 1 \times 1$, where $HW$ is the number of the kernels, $C$ is the number of channels for each kernel and $1 \times 1$ is the size of each kernel. In $K^t$ and $K^b$, each kernel corresponds to a weighted feature at a special position in $F^0$.

With $K^t$ and $K^b$, the matching process can be implemented by the convolution operations as follows:

$$S^t = F * K^t, S^t \in \mathbb{R}^{1 \times (HW) \times H \times W} \quad (3)$$

$$S^b = F * K^b, S^b \in \mathbb{R}^{1 \times (HW) \times H \times W} \quad (4)$$

Each element in $S^t$ and $S^b$ is the cosine distance of two directional feature vectors:

$$S^t_{1,k,h,w} = \sum_{c=1}^{C} \left( F_{1,c,h,w} \cdot K^t_{k,c,1,1} \right) \quad (5)$$

$$S^b_{1,k,h,w} = \sum_{c=1}^{C} \left( F_{1,c,h,w} \cdot K^b_{k,c,1,1} \right) \quad (6)$$

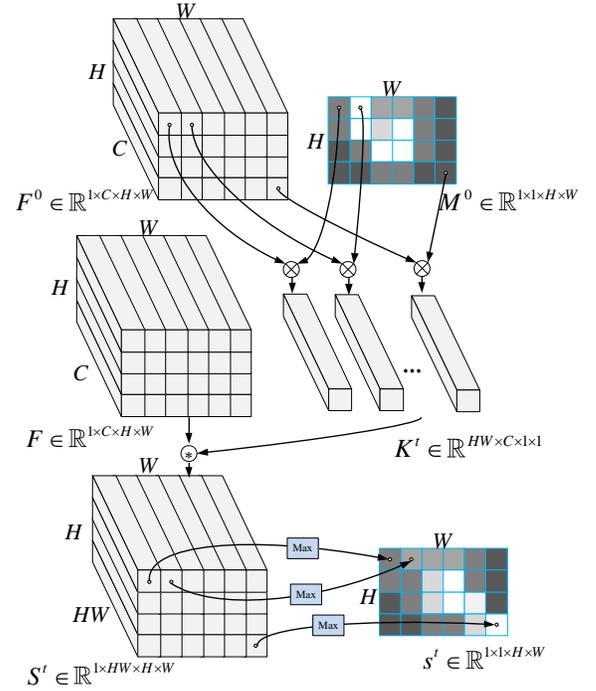

Fig. 2. The process of global directional matching for the target. $F^0 \in \mathbb{R}^{1 \times C \times H \times W}$ represents the normalized feature map of the first frame. $M^0 \in \mathbb{R}^{1 \times 1 \times H \times W}$ represents the resized mask of the first frame. $K^t \in \mathbb{R}^{HW \times C \times 1 \times 1}$ represents the generated convolutional kernels. $F \in \mathbb{R}^{1 \times C \times H \times W}$ represents the normalized feature maps of the input frame. $S^t \in \mathbb{R}^{1 \times HW \times H \times W}$ represents the output of the convolutional operation between $F \in \mathbb{R}^{1 \times C \times H \times W}$ and $K^t \in \mathbb{R}^{HW \times C \times 1 \times 1}$ with stride 1. $s^t \in \mathbb{R}^{1 \times 1 \times H \times W}$ represents the results of global directional matching by taking the maximum value in each channel of $S^t \in \mathbb{R}^{1 \times (HW) \times H \times W}$.

The final outputs $s^t \in \mathbb{R}^{1 \times 1 \times H \times W}$ and $s^b \in \mathbb{R}^{1 \times 1 \times H \times W}$ of global directional matching are obtained by taking the maximum value in each channel of $S^t$ and $S^b$:

$$s^t_{1,1,h,w} = \max_c S^t_{1,c,h,w} \tag{7}$$

$$s^b_{1,1,h,w} = \max_c S^b_{1,c,h,w} \tag{8}$$

We directly use $s^t_{1,1,h,w}$ and $s^b_{1,1,h,w}$ as the outputs of our global directional matching and feed them into the convolution layers of the fusion module because our global directional matching is only used as an internal guidance of the convolutional network. Of course, for visualization, a simple soft-max operation can be applied on $s^t_{1,1,h,w}$ and $s^b_{1,1,h,w}$ to obtain the probabilities of the target and background.

*C. Directional statistics-based appearance learning*

Our directional statistics-based appearance module is to learn a vMF-based generative appearance model of the video content in a spherical embedding space. It aims to dynamically update the cues of the target and background with progressing of video frames. Given a new frame, our model will output pixel-embedding's posterior class probabilities for foreground/background discrimination. The entire procedure of our directional appearance model learning is summarized in Algorithm 1.

**Directional appearance model.** The von Mises–Fisher (vMF) distribution [42] is widely used to model directional statistics. It is a probability distribution on the ($p$-1)-dimensional sphere in $\mathbb{R}^p$. The vMF distribution of a $p$-dimensional random directional variable $r$ with unit $L_2$-norm is $f_p(r; \mu, \kappa) = Z_p(\kappa)\exp(\kappa \mu^T r)$, where $\kappa \geq 0$, $\|\mu\|=1$ and the normalization coefficient $Z_p(\kappa) = \frac{\kappa^{p/2-1}}{(2\pi)^{p/2} I_{p/2-1}(\kappa)}$. Here $I_v$ is the modified Bessel function of the first type with order $v$ and $p$ is the dimensionality of $r$. There are two parameters in a vMF distribution: the mean direction $\mu$ and the concentration parameter $\kappa$ that characterizes the tightness of the distribution around the mean direction $\mu$.

Denote by $F$ the normalized feature maps extracted from the current frame, and by $r_l$ the directional feature vector at spatial location $l$ of $F$. We model the extracted directional vectors as i.i.d samples drawn from the underlying distribution

$$p(r_l) = \sum_{k=1}^{K} p(z_l = k) p(r_l | z_l = k) \tag{9}$$

Each class-conditional density is a vMF distribution with the mean direction $\mu_k$ and the concentration parameter $\kappa_k$:

$$p(r_l | z_l = k) = f_p(r_l; \mu_k, \kappa_k) \tag{10}$$

The observed directional feature $r_l$ is assigned to a specific component $z_l = k$ based on its discrete probability. A uniform prior $p(z_l = k) = 1/K$ is used for the random variable $z_l$, where $K$ is the number of components.

Given the mixture model parameters $\mu_k, \kappa_k$, $k \in \{0,1,\ldots,K-1\}$, the component posteriors can be calculated as follows:

$$p(z_l = k | r_l, \mu, \kappa) = \frac{p(z_l = k) p(r_l | z_l = k)}{\sum_k p(z_l = k) p(r_l | z_l = k)} \tag{11}$$

Substituting Eq. (10) into Eq. (11), we have:

$$p(z_l = k | r_l, \mu, \kappa) = \frac{Z_p(\kappa_k) \exp(\kappa_k (\mu_k)^T r_l)}{\sum_k Z_p(\kappa_k) \exp(\kappa_k (\mu_k)^T r_l)} \tag{12}$$

In order to avoid calculating the complex normalization constant $Z_p(\kappa)$, we set the parameter $\kappa_k$ of different components to equal, so Eq. (12) can be simplified as follows:

$$p(z_l = k | r_l, \mu, \kappa) = \frac{\exp(\kappa (\mu_k)^T r_l)}{\sum_k \exp(\kappa (\mu_k)^T r_l)} \tag{13}$$

In practice, we use four vMF distributions, where the components 0 and 2 model the directional features of background and components 1 and 3 model the directional features of the target. Similar to the strategy adopted in A-GAME [12], we use one pair of base components to predict background $k = 0$ and foreground $k = 1$, and another pair of supplementary components to predict the background $k = 2$ and foreground $k = 3$, which are incorrectly classified by the base components.

As done in the global directional matching module described in Section III.B, we use $s_{lk} = \kappa (\mu_k)^T r_l, k \in \{0,1,2,3\}$ as the output of each component and feed them into the convolution layers in the fusion module.

**Model parameter estimation**. The parameters $\mu_k$ and $\kappa$ need to be estimated in the posterior class probabilities shown in Eq. (13). For semi-supervised segmentation, an initial segmentation mask defining the target in the first frame is given. So the parameters of our directional appearance model are estimated from the extracted directional features and the given mask in the first frame. In subsequent frames, the network predictions are used as soft class labels to estimate these parameters dynamically. The parameter $\kappa$ is a scalar, and we set it as a trainable parameter in our network.

Given the set of directional features $\{r_l\}_l$, the parameter $\mu$ can be estimated as $\mu^i_k = \sum_l r^i_l / \|\sum_l r^i_l\|$ by the maximum likelihood estimation [42]. For the first frame, the parameter $\mu^0_k$ is estimated as $\mu^0_k = \sum_l \alpha^0_{l,k} r^0_l / \|\sum_l \alpha^0_{l,k} r^0_l\|$, where $\alpha^0_{l,k} \in \{0,1\}$ is the soft class labels, denoting the level of assigning the directional feature $r^i_l$ to a specific component $k$. In subsequent frames, the parameter $\mu^i_k$ is updated with the previous estimates using a learning rate $\lambda$:

$$\mu^i_k = (1-\lambda)\mu^{i-1}_k + \lambda \overline{\mu}^i_k \tag{14}$$

where $\overline{\mu}^i_k = \sum_l \alpha^i_{l,k} r^i_l / \|\sum_l \alpha^i_{l,k} r^i_l\|$.

For the pair of base components $k \in \{0, 1\}$, $\alpha^i_{l,k}$ is calculated as follows:

$$\begin{cases} \alpha^i_{l,0} = 1 - y_l(I^i, \mu^{i-1}_k, \kappa, \Phi) \\ \alpha^i_{l,1} = y_l(I^i, \mu^{i-1}_k, \kappa, \Phi) \end{cases} \tag{15}$$

where $y_l(I^i, \mu^{i-1}_k, \kappa^{i-1}, \Phi) \in [0,1]$ is the probability of the target

**Algorithm 1:** The inference and update process of directional appearance model

**Inference process:** Output the dynamically updated cues $s_{lk}^{i}, k \in \{0,1,2,3\}$ based on the appearance model parameters $\boldsymbol{\mu}_{k}^{i}$ and $\kappa$, and the input directional feature $r_{p}^{i}$.

**update process:** Output the updated appearance model parameters $\mu_{k}^{i}$ based on the probabilities $y_{l}\left(I^{i}, \mu_{k}^{i-1}, \kappa, \Phi\right) \in [0,1]$ of the target predicted by the network.

---

Inference ($r_p^i$, $\mu_k^{i-1}$, $\kappa$):

    for $k$ = 0, 1, 2, 3: compute $s_{lk}^{i} = \kappa \left(\boldsymbol{\mu}_{k}^{i-1}\right)^{T} r_{l}^{i}$

    **return** $s_{lk}^{i}, k \in \{0,1,2,3\}$

Update ($r_p^i$, $\mu_k^{i-1}$, $\kappa$, $y_l$):

    for $k$=0, 1: compute $\alpha_{l,k}^{i}$ from (15)

    for $k$=0, 1: compute $\mu_{k}^{i}$ from (14)

    for $k$=0, 1: compute

$$p\left(z_{l}^{i}=k \mid r_{l}^{i}, \boldsymbol{\mu}_{k \in\{0,1\}}^{i}, \kappa\right)=\frac{\exp(\kappa(\boldsymbol{\mu}_{k}^{i})^{T} r_{l}^{i})}{\sum_{k=0}^{1}\exp(\kappa(\boldsymbol{\mu}_{k}^{i})^{T} r_{l}^{i})}$$

    for $k$=2, 3: compute $\alpha_{l,k}^{i}$ from (16)

    for $k$=2, 3: compute $\mu_{k}^{i}$ from (14)

    **return** $\mu_{k}^{i}, k \in \{0,1,2,3\}$

---

predicted by the network, $I^i$ is the input image of the $i$th frame, $\Phi$ is the network parameters, $\mu_k^{i-1}$ and $\kappa$ are currently estimated appearance model parameters. In particular, $\alpha_{l,k}^{0} \in \{0,1\}$ can be directly obtained from the first frame $i$=0 since the pixel labels of the target and background are given. For the pair of supplementary components $k \in \{2, 3\}$, $\alpha_{l,k}^{i}$ is calculated as follows:

$$\begin{cases} \alpha_{l,2}^{i} = \max\left(0, \alpha_{l,0}^{i} - p\left(z_{l}^{i}=0 \mid r_{l}^{i}, \boldsymbol{\mu}_{k \in\{0,1\}}^{i}, \kappa\right)\right) \\ \alpha_{l,3}^{i} = \max\left(0, \alpha_{l,1}^{i} - p\left(z_{l}^{i}=1 \mid r_{l}^{i}, \boldsymbol{\mu}_{k \in\{0,1\}}^{i}, \kappa\right)\right) \end{cases} \quad (16)$$

where $p\left(z_{l}^{i}=k \mid r_{l}^{i}, \boldsymbol{\mu}_{k \in\{0,1\}}^{i}, \kappa\right)$ denotes the posteriors shown in Eq. (13), evaluated using only the base components. The $\alpha_{l,2}^{i}$ and $\alpha_{l,3}^{i}$ enforce the supplementary components to concentrate on directional features which are incorrectly classified by the posteriors $p\left(z_{l}^{i}=k \mid r_{l}^{i}, \boldsymbol{\mu}_{k \in\{0,1\}}^{i}, \kappa\right)=\alpha_{l,k}^{i}$, $k$=0, 1.

## IV. EXPERIMENTS

### A. Implementation Details

**The network settings.** We use ResNet50 or ResNet101 [43] as the backbone feature extractor, and implement layer5 with dilated convolutions [44], producing features with a stride of 16. In the global matching module, we add an embedding layer which consists of a 1 ×1 convolution with stride 1, reducing the channels of the backbone feature maps from 2048 to 512. Then we normalize the embedding features to obtain the directional feature maps with 512 channels. The global matching process is performed on these directional features with 512 dimensions. In our directional appearance learning module, a 1 × 1 convolution with stride 1 and the normalization operation are applied on the backbone feature maps to obtain directional feature maps with 512 channels. The outputs of mask propagation, global matching module and directional appearance module are concatenated and fed into the fusion module. The cross-entropy losses of the predicted coarse segmentation and final segmentation are summed and minimized by Adam [46].

**The training details.** The backbone of our model is initialized by the weights trained on ImageNet [45] and only the layer5 in the backbone and other modules are trained. Then we train the proposed network on the DAVIS 2017 [14] training set, the YouTube-VOS [15] training set and synthetic dataset. The DAVIS 2017 training set includes 60 videos and each video contains one or several annotated objects. The YouTube-VOS training set is composed of 3,471 videos, each of which is labeled for every five frames. Each video in YouTube-VOS training set includes one or several annotated objects. As in [12][25], our synthetic dataset is generated by pasting segmented targets on background images. Specifically, we randomly select 1 to 5 segmented objects from COCO [47] and a video from ILSVRC2017_VID [45], and then paste the selected objects after an affine transformation onto each frame of the selected video.

Our model is first trained with 100 epochs using videos with a lower resolution of 240 × 432. The initial value of the parameter $\kappa$ is set to 30. The batch size is set to 32 which includes 4 video snippets with 8 frames in each. A learning rate of $10^{-4}$, an exponential learning rate decay of 0.95 per epoch, and a weight decay of $10^{-5}$ are used. Our model is then trained with another 120 epochs on videos with a higher resolution of 480 × 864. The batch size is set to 24 which includes 2 video snippets with 12 frames in each. A learning rate of $10^{-5}$, an exponential learning rate decay of 0.985 per epoch, and a weight decay of $10^{-6}$ are used.

### B. Experimental Results

We compare our proposed DDEAL with recent state-of-the-art methods on three datasets, including DAVIS 2016 [13], DAVIS 2017 [14] and the recent large-scale YouTube-VOS dataset [15]. Our approach is implemented in PyTorch and trained on a single NVIDIA TITAN Xp GPU. DDEAL runs in real time with 25 fps (frames per second). Our code and trained networks are available at https://github.com/YingjieYin/Directional-Deep-Embedding-and-Appearance-Learning-for-Fast-Video-Object-Segmentation.

**Evaluation on DAVIS2016.** The DAVIS-2016 database is a video dataset for single object segmentation and it consists of 50 video sequences. We use two important measures, the mean Jaccard index $\mathcal{J}$, i.e. intersection-over-union (IoU), and the mean contour accuracy $\mathcal{F}$ to evaluate the segmentation performance. All evaluation results are computed on DAVIS-2016's validation set which includes 20 video sequences.

We compare our proposed DDEAL with state-of-the-art fine-tuning-dependent and fine-tuning-free VOS methods. The compared fine-tuning-dependent methods include OSVOS-S [17], OnAVOS [18], MGCRN [49], CINM [48], Lucid [21], OSVOS [16], MSK [20], S2S [15] and SegFlow [19]. The compared fine-tuning-free methods include A-GAME [12], RGMP [25], FEELVOS [11], FAVOS [24], VM [26], PML [10], OSMN [27], CTN [28] and VPN [29]. Table I shows the quantitative evaluation results of all competing methods. Res50 and Res101 in Table I represent our DDEAL using ResNet50 and ResNet101, respectively, as the backbone. In terms of the $\mathcal{J}\&\mathcal{F}$ Mean, DDEAL-Res101 ranks the third place and is only 0.1% and 1.2% lower than OnAVOS and OSVOS-S, respectively. However, OnAVOS and OSVOS-S rely on fine-tuning a network by online learning on the first frame or previous frame of each test video, which results in slow running speeds. DDEAL-Res101 runs at 25 fps on average and it is 325 times faster than OnAVOS and 112 times faster than OSVOS-S. In addition, as we will see later, OnAVOS and OSVOS-S do not generalize well to the larger and more diverse DAVIS2017 and YouTube-VOS datasets. Among the fine-tuning-free methods, our method significantly outperforms all the other ones in both accuracy and speed. As shown in Table I, the $\mathcal{J}\&\mathcal{F}$ Mean of DDEAL-Res101 is 3.3%, 3.6% and 3.7% higher than A-GAME, RGMP and FEELVOS, respectively. Meanwhile, DDEAL-Res101 is 1.7, 3.3 and 11.3 times faster than A-GAME, RGMP and FEELVOS, respectively. In terms of the $\mathcal{J}\&\mathcal{F}$ Mean, our faster version DDEAL-Res50 ranks the fourth place, what's more, it runs the fastest with 31 fps on average. Fig. 3 shows different method's $\mathcal{J}\&\mathcal{F}$ Mean with respect to their time to process one frame in DAVIS2016.

TABLE I
QUANTITATIVE RESULTS ON DAVIS2016 VALIDATION SET. FOR EACH METHOD, WE REPORT WHETHER IT EMPLOYS ONLINE FINE-TUNING (O-FT), THE MEAN

TABLE I
INTERSECTION OVER UNION $\mathcal{J}$, THE CONTOUR ACCURACY $\mathcal{F}$, $\mathcal{J}\&\mathcal{F}$ MEAN
AND THE RUNNING TIME PER FRAME IN SECONDS. THE BEST THREE RESULTS
FOR EACH MEASURE ARE HIGHLIGHTED IN BOLD.

| Method | | O-Ft | $\mathcal{J}$ (%) | $\mathcal{F}$ (%) | $\mathcal{J}\&\mathcal{F}$ (%) | Time |
|---|---|---|---|---|---|---|
| OSVOS-S [17] | | ✓ | **85.6** | **87.5** | **86.6** | 4.5s |
| OnAVOS [18] | | ✓ | **86.1** | 84.9 | **85.5** | 13s |
| MGCRN [49] | | ✓ | 84.4 | **85.7** | 85.1 | 0.73s |
| CINM [48] | | ✓ | 83.4 | 85.0 | 84.2 | >30s |
| Lucid [21] | | ✓ | 83.9 | 82.0 | 83.0 | >100s |
| OSVOS [16] | | ✓ | 79.8 | 80.6 | 80.2 | 9s |
| MSK [20] | | ✓ | 79.7 | 75.4 | 77.6 | 12s |
| S2S [15] | | ✓ | 79.1 | - | - | 9s |
| SegFlow [19] | | ✓ | 74.8 | 76.0 | 75.4 | 7.9s |
| A-GAME [12] | | ✗ | 82.0 | 82.2 | 82.1 | **0.07s** |
| RGMP [25] | | ✗ | 81.5 | 82.0 | 81.8 | 0.13s |
| FEELVOS [11] | | ✗ | 81.1 | 82.2 | 81.7 | 0.45 |
| FAVOS [24] | | ✗ | 82.4 | 79.5 | 81.0 | 1.80s |
| VM [26] | | ✗ | 81.0 | 80.8 | 80.9 | 0.32s |
| PML [10] | | ✗ | 75.5 | 79.3 | 77.4 | 0.28s |
| OSMN [27] | | ✗ | 74.0 | 72.9 | 73.5 | 0.14s |
| CTN [28] | | ✗ | 73.5 | 69.3 | 71.4 | 1.30s |
| VPN [29] | | ✗ | 70.2 | 65.5 | 67.9 | 0.63s |
| DDEAL(ours) | Res50 | ✗ | 85.0 | 84.6 | 84.8 | **0.03s** |
| | Res101 | ✗ | 85.1 | **85.7** | 85.4 | **0.04s** |

The A-GAME [12] learns target appearance described by the mixture of Gaussians in the Euclidean space to achieve fine-tuning-free VOS. In order to show the reason that our directional appearance model described by the mixture of vMF on a hyper sphere can more robustly represent the foreground and background's appearances than the appearance model described in A-GAME [12], we visualize the learned cues from our DDEAL and A-GAME in Fig.4. Fig.4.(a) shows the probability map obtained from DDEAL's global direction matching module described in Section III. B. Fig.4.(b) and Fig.4.(c) shows the probability maps generated by DDEAL's base and supplementary components, respectively, described in Section III. C. Fig.4.(d) and Fig.4.(e) shows the probability maps generated by A-GAME's base and supplementary components, respectively. Compared Fig.4.(b) with Fig.4.(d), we can see that the base pixel-embedding's posterior class probabilities for foreground/background discrimination in our DDEAL are more accurate than A-GAME's. Otherwise, as illustrated in Fig.4.(b) and Fig.4.(c), the failure cues (highlighted by rectangles) in the base components in our DDEAL can be better predicted by the corresponding supplementary components.

**Evaluation on DAVIS2017.** The DAVIS-2017 validation set [14] consists of 30 challenging videos, each of which has one or multiple target objects. The experimental results are shown in Table II. The intersection-over-union $\mathcal{J}$ and contour accuracy $\mathcal{F}$ are used as metrics for performance evaluation on this dataset. We compare our proposed DDEAL with recent state-of-the-art methods, including CINM [48], OSVOS-S [17], Lucid [21], OnAVOS [18], OSVOS [16], FEELVOS [11], A-GAME [12], RGMP [25], VM [26], FAVOS [24] and OSMN [27]. As shown in Table II, DDEAL-Res101 and DDEAL-Res50 achieve the best results in all measures. The $\mathcal{J}$&$\mathcal{F}$ Mean of DDEAL-Res101 is 3.3%, 4.2% and 4.8% higher than the third best method FEELVOS [11], the fourth best method CINM [48] and the fifth best method A-GAME [12], respectively. The $\mathcal{J}$&$\mathcal{F}$ Mean of our faster version DDEAL-Res50 is also 3.2%, 4.1% and 4.7% higher than FEELVOS [11], CINM [48] and A-GAME [12], respectively.

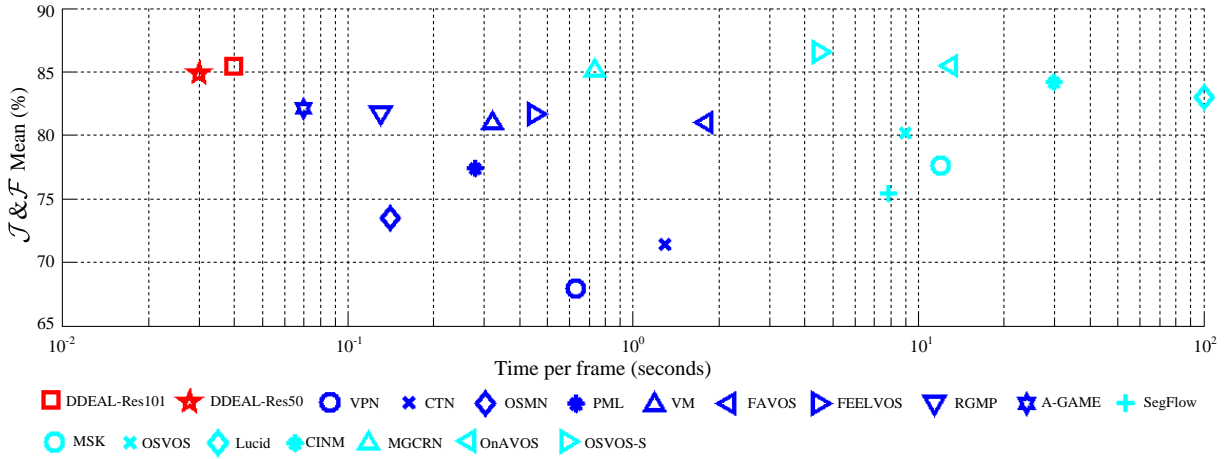

Fig. 3. Performance versus timing in DAVIS-2016: $\mathcal{J}\&\mathcal{F}$ Mean with respect to their time to process one frame. The better methods are located at the upper-left corner.

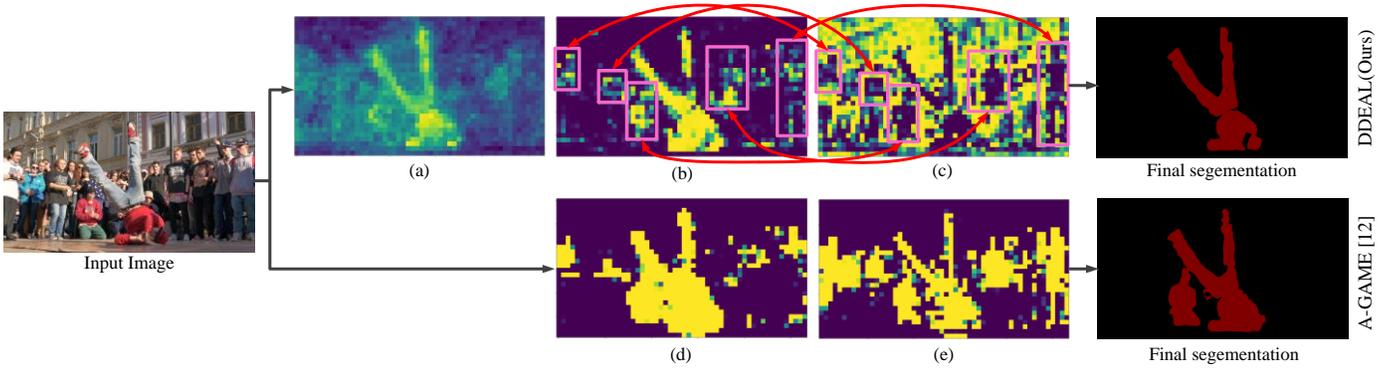

Fig. 4. Comparison between our proposed DDEAL (top row) and the recently proposed A-GAME [12] (bottom row). For DDEAL: (a) shows the probability map obtained from global direction matching; (b) shows the probability map generated by the base components of the directional statistics-based appearance model; and (c) shows the probability map generated by the supplementary components of the appearance model. For A-GAME: (d) shows the probability map generated by the base components of the Gaussian-distribution-based appearance model; and (e) shows the probability map generated by the supplementary components of the appearance model. For visualization, the probability maps are resized to match the size of the original image. One can see that the failure cues (highlighted by rectangles) in the base components can be better predicted by the corresponding supplementary components in our DDEAL.

TABLE II
QUANTITATIVE RESULTS ON DAVIS2017 VALIDATION SET. FOR EACH METHOD, WE REPORT WHETHER IT EMPLOYS ONLINE FINE-TUNING (O-FT), THE MEAN INTERSECTION OVER UNION $\mathcal{J}$, THE CONTOUR ACCURACY $\mathcal{F}$ AND $\mathcal{J}$ & $\mathcal{F}$ MEAN. THE BEST THREE RESULTS FOR EACH MEASURE ARE HIGHLIGHTED IN BOLD.

| Method | O-Ft | $\mathcal{J}$ (%) | $\mathcal{F}$ (%) | $\mathcal{J}$ & $\mathcal{F}$ (%) |
|---|---|---|---|---|
| CINM [48] | ✓ | 67.2 | **74.0** | 70.6 |
| OSVOS-S [17] | ✓ | 64.7 | 71.3 | 68.0 |
| Lucid [21] | ✓ | 63.4 | 69.9 | 66.6 |
| OnAVOS [18] | ✓ | 61.6 | 69.1 | 65.4 |
| OSVOS [16] | ✓ | 56.6 | 63.9 | 60.3 |
| FEELVOS [11] | ✗ | **69.1** | **74.0** | **71.5** |
| A-GAME [12] | ✗ | 67.2 | 72.7 | 70.0 |
| RGMP [25] | ✗ | 64.8 | 68.6 | 66.7 |
| VM [26] | ✗ | 56.5 | 68.2 | 62.4 |
| FAVOS [24] | ✗ | 54.6 | 61.8 | 58.2 |
| OSMN [27] | ✗ | 52.5 | 57.1 | 54.8 |
| DDEAL(ours) Res50 | ✗ | **72.1** | **77.4** | **74.7** |
| DDEAL(ours) Res101 | ✗ | **72.0** | **77.6** | **74.8** |

TABLE III
QUANTITATIVE RESULTS ON YOUTUBE-VOS VALIDATION SET. OUR APPROACH OBTAINS THE BEST RESULTS IN ALL MEASURES THOUGH IT DOES NOT PERFORM ANY ONLINE FINE-TUNING (O-FT). THE BEST THREE RESULTS FOR EACH MEASURE ARE HIGHLIGHTED IN BOLD.

| Method | O-Ft | Seen $\mathcal{J}$ (%) | Seen $\mathcal{F}$ (%) | Unseen $\mathcal{J}$ (%) | Unseen $\mathcal{F}$ (%) | Overall $\mathcal{G}$ (%) |
|---|---|---|---|---|---|---|
| S2S [15] | ✓ | **71.0** | 70.0 | 55.5 | **61.2** | 64.4 |
| OnAVOS [18] | ✓ | 60.1 | 62.7 | 46.1 | 51.4 | 55.2 |
| OSVOS [16] | ✓ | 59.8 | 60.5 | 54.2 | 60.7 | 58.8 |
| MSK [20] | ✓ | 59.9 | - | 53.1 | - | 53.1 |
| A-GAME [12] | ✗ | 66.9 | - | **61.2** | - | **66.0** |
| RGMP [25] | ✗ | 59.5 | - | 53.8 | - | 53.8 |
| OSMN [27] | ✗ | 60.0 | 60.1 | 51.2 | 44.0 | 51.2 |
| DDEAL(ours) Res50 | ✗ | **72.5** | **75.8** | **63.4** | **70.4** | **70.5** |
| DDEAL(ours) Res101 | ✗ | **73.7** | **77.1** | **63.9** | **70.7** | **71.3** |

TABLE V
ABLATION STUDY ON DAVIS 2017 AND YOUTUBE-VOS VALIDATION SET. GDMM DENOTES THE GLOBAL DIRECTIONAL MATCHING MODULE, AND DALM DENOTES THE DIRECTIONAL APPEARANCE LEARNING MODULE. BC AND SC DENOTE THE BASE COMPONENTS AND SUPPLEMENTARY COMPONENTS OF DALM, RESPECTIVELY.

| DDEAL (Res101) | | | DAVIS 2017 validation set | | | YouTube-VOS validation set | | | | |
|---|---|---|---|---|---|---|---|---|---|---|
| GDMM | DALM BC | DALM SC | $\mathcal{J}$ (%) | $\mathcal{F}$ (%) | $\mathcal{J}$ & $\mathcal{F}$ (%) | Seen $\mathcal{J}$ (%) | Seen $\mathcal{F}$ (%) | Unseen $\mathcal{J}$ (%) | Unseen $\mathcal{F}$ (%) | Overall $\mathcal{G}$ (%) |
| 1 ✓ | ✓ | ✓ | 72.0 | 77.6 | 74.8 | 73.7 | 77.1 | 63.9 | 73.7 | 71.3 |
| 2 ✓ | | | 70.1 | 74.1 | 72.1 | 72.2 | 74.9 | 60.4 | 72.2 | 68.2 |
| 3 | ✓ | ✓ | 68.3 | 73.8 | 71.1 | 70.0 | 72.5 | 60.0 | 70.0 | 67.3 |
| 4 ✓ | ✓ | | 72.0 | 77.6 | 74.8 | 72.6 | 75.8 | 62.9 | 72.6 | 70.1 |
| 5 | | ✓ | 65.6 | 71.4 | 68.5 | 69.0 | 71.6 | 57.3 | 69.0 | 65.1 |

**Evaluation on YouTube-VOS.** The validation set of YouTube-VOS comprises 474 videos labeled with one or multiple objects. Our results are obtained through an official online evaluation system since the ground-truth masks of the validation set are withheld. Intersection-over-union $\mathcal{J}$ and contour accuracy $\mathcal{F}$ are separately calculated for seen and unseen classes by the online evaluation system, resulting in four performance measures. The average of all the four measures is calculated as the overall performance $\mathcal{G}$. We compare our proposed DDEAL with recent state-of-the-art methods S2S[15], OnAVOS[18], OSVOS[16], MSK[20], A-GAME [12], RGMP [25] and OSMN[27]. As shown in Table III, DDEAL-Res101 and DDEAL-Res50 achieve the best results in all measures though they do not perform any online fine-tuning. In particular, the overall performance $\mathcal{G}$ of DDEAL-Res101 and DDEAL-Res50 are 5.3% and 4.5% higher than the second best method A-GAME [12].

## C. Ablation Study

In Table V, we analyze the effect of DDEAL's individual components on the DAVIS 2017 and YouTube-VOS validation set. Line 1 shows the results of DDEAL with a ResNet101 backbone using both the global directional matching module (GDMM) and the directional appearance learning module (DALM). In line 2, we disable the DALM, and the $\mathcal{J}\&\mathcal{F}$ Mean on the DAVIS 2017 and the overall performance $\mathcal{G}$ on YouTube-VOS are dropped by 2.7% and 3.1%, respectively. It demonstrates that the dynamically updated cues provided by GDMM are important for DDEAL to achieve good results. In line 3, the GDMM is disabled and the $\mathcal{J}\&\mathcal{F}$ Mean on the DAVIS 2017 and the overall performance $\mathcal{G}$ on YouTube-VOS are dropped by 3.7% and 4.0%, respectively, which demonstrates that the static cues provided by GDMM are also important for DDEAL to achieve good results. In line 4, the supplementary components (SC) of DALM are disabled, and the overall performance $\mathcal{G}$ on YouTube-VOS is dropped by 1.2%, which shows that the multi-modal generative model helps to improve performance. In line 5, only the base components (BC) of DALM are enabled in DDEAL; in this case, the results deteriorate to 68.5% and 65.1% on the DAVIS 2017 and YouTube-VOS, respectively, showing that GDMM and multi-modal generative model are key factors for our proposed DDEAL.

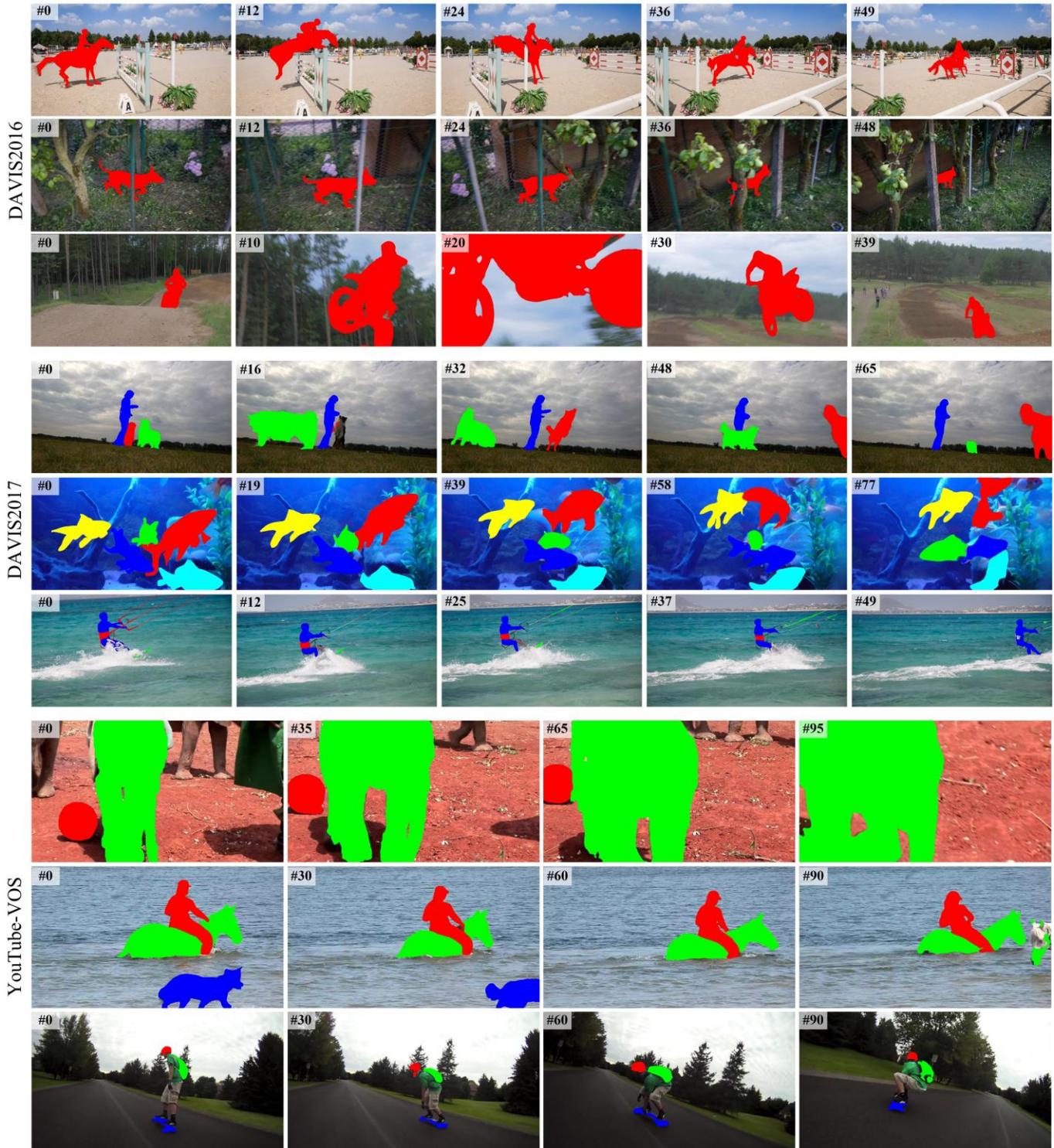

Fig. 5. Some Qualitative results on the DAVIS2016, DAVIS2017 and YouTube-VOS validation sets.

*D. Qualitative Results*

Some qualitative results of DDEAL on the DAVIS 2016 validation set, the DAVIS 2017 validation set and the YouTube-VOS validation set are shown in Fig. 5. One can see that in many cases DDEAL is able to segment objects accurately, even in difficult cases such as large scale changing in the horsejump-high (first row), the motocross-jump sequence (third row) and the ball-elephant sequence (seventh row), occlusion in the libby sequence (second row), similar appearances in the dogs-jump (forth row) and the gold-fishsequence (fifth row), and large motion in the hat-bag-skateboard sequence (last row). In the challenging kite-surf sequence (sixth row), DDEAL fails to segment the lines accurately. This is probably because the lines are so thin that they are drowned in the background during the convolutional operations and the direction-based embedding feature matching and appearance model fails to describe them.

## V. CONCLUSIONS

We presented a directional deep embedding and appearance learning (DDEAL) approach for fast and accurate semi-supervised VOS. DDEAL exploits directional cues for accurate segmentation by two key modules: global directional matching module and directional appearance learning module. The former extracts static cues from the labeled first frame by pixel-wise embedding, while the later learns dynamic cues from the subsequent frames by a vMF-based generative appearance model. Thanks to the directional features, both the modules can be implemented efficiently and DDEAL runs at a speed of 25 fps. The effectiveness of DDEAL was also validated on benchmark datasets, where we achieved leading VOS performance on DAVIS 2017 and the large scale YouTube-VOS datasets. As a fine-tuning-free method, DDEAL provides a practical solution to VOS in real applications.

# Supplementary File to
# "Directional Deep Embedding and Appearance Learning for Fast Video Object Segmentation"

In this supplementary file, we provide qualitative comparison between our DDEAL method and some representative and state-of-the-art VOS methods, including A-GAME [1], FEELVOS [3] and RGMP [2]. Figs. 1-3, Figs. 4-9 and Figs.10-15 show the qualitative results on the DAVIS-2016, DAVIS-2017 and YouTube-VOS validation sets, respectively. (Since the visual results of FEELVOS and RGMP on YouTube-VOS are not available, we only compared with A-GAME on YouTube-VOS.) It can be seen that DDEAL achieves more accurate segmentation performance than the competing methods on both single-object and multi-object tasks. Demo videos are also included in our uploaded supplementary materials.

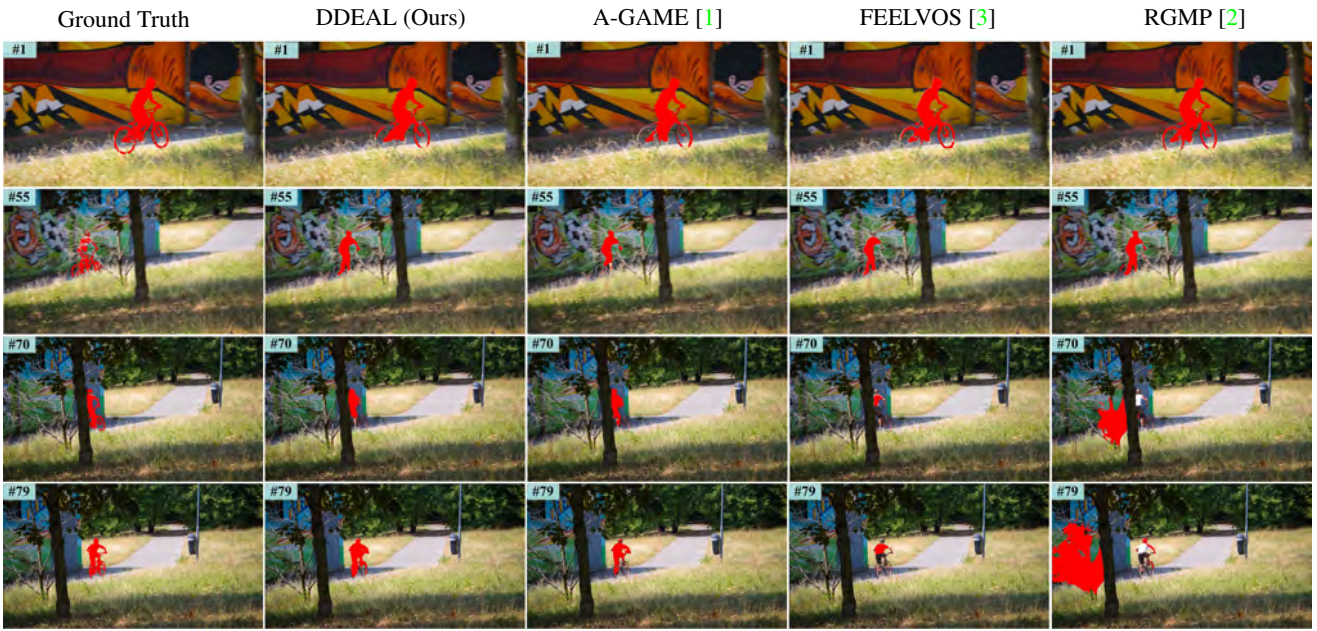

Figure 1: Qualitative results of different methods on the bmx-trees sequence of the DAVIS-2016 validation set.

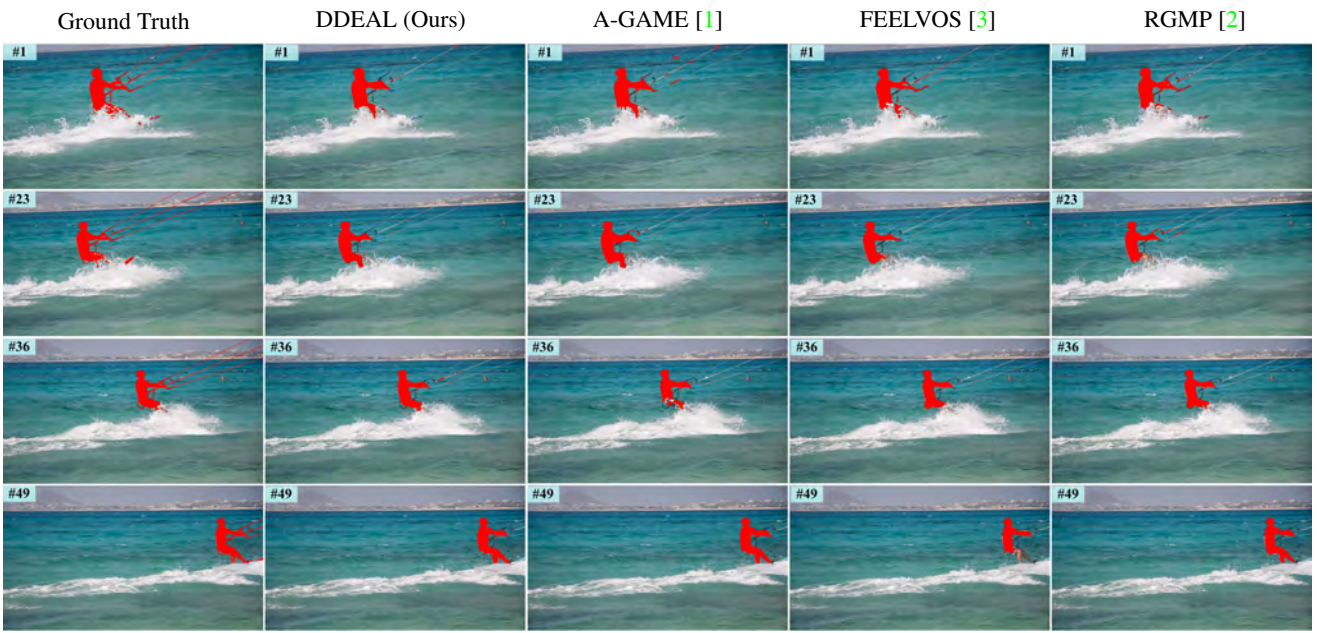

Figure 2: Qualitative results of different methods on the kite-surf sequence of the DAVIS-2016 validation set.

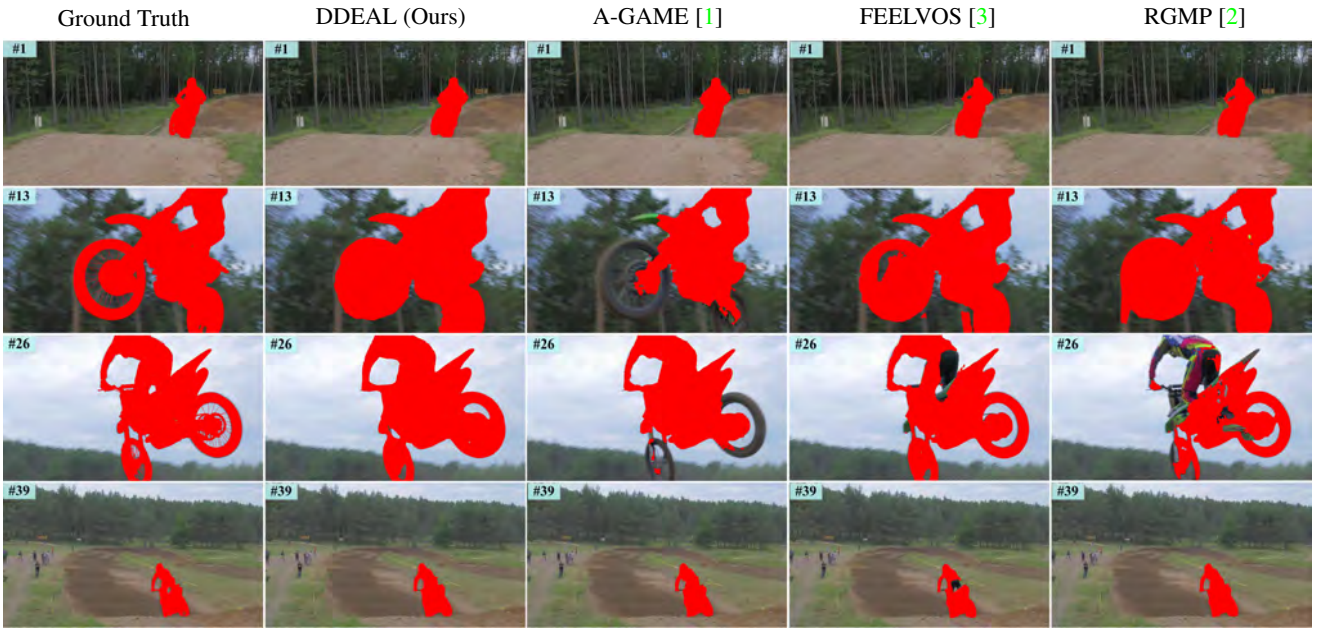

Figure 3: Qualitative results of different methods on the motocross-jump sequence of the DAVIS-2016 validation set.

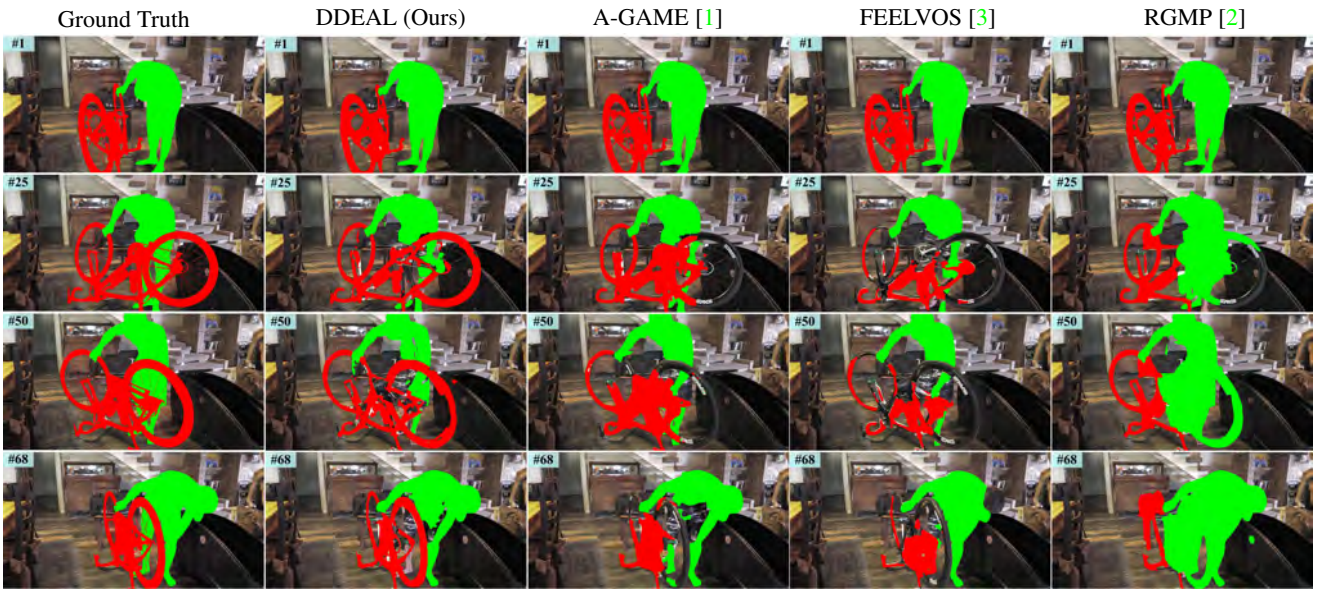

Figure 4: Qualitative results of different methods on the bike-packing sequence of the DAVIS-2017 validation set.

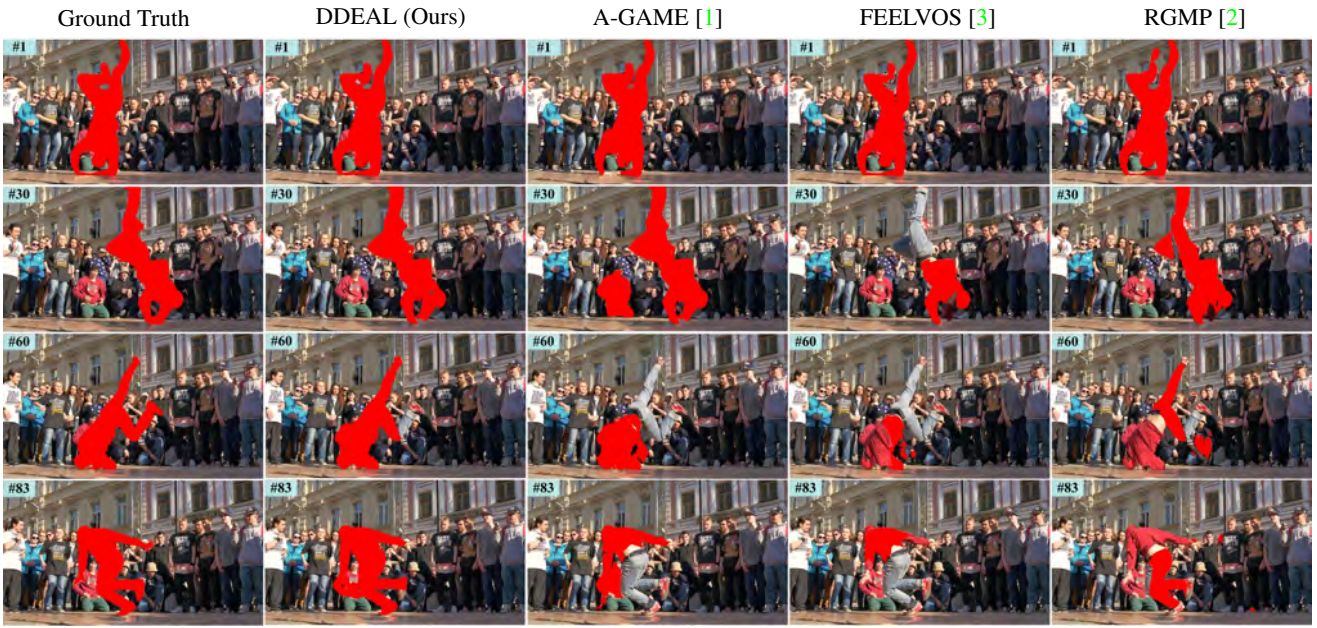

Figure 5: Qualitative results of different methods on the breakdance sequence of the DAVIS-2017 validation set.

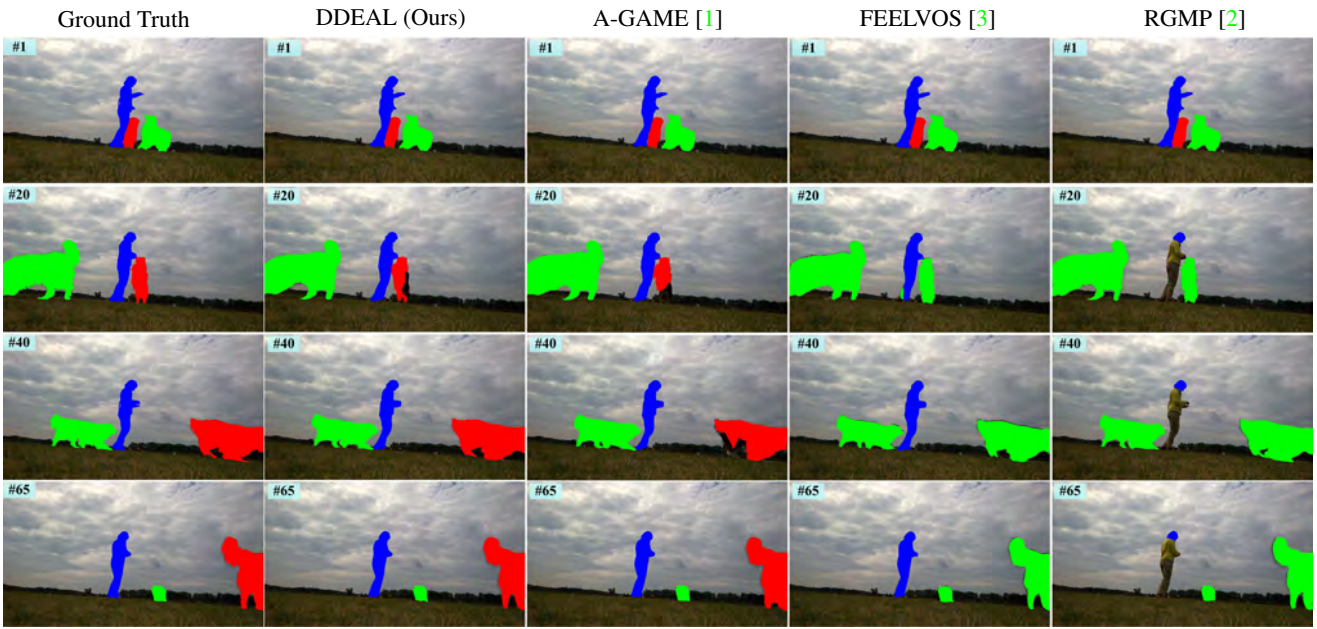

Figure 6: Qualitative results of different methods on the dogs-jump sequence of the DAVIS-2017 validation set.

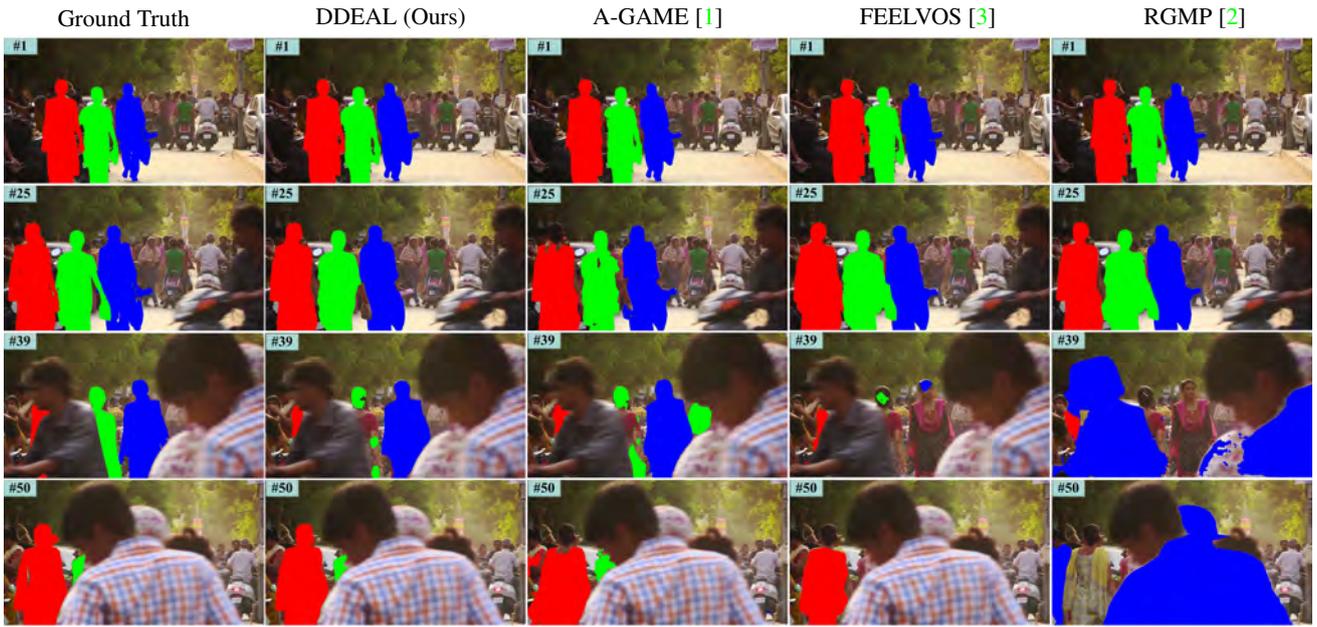

Figure 7: Qualitative results of different methods on the india sequence of the DAVIS-2017 validation set.

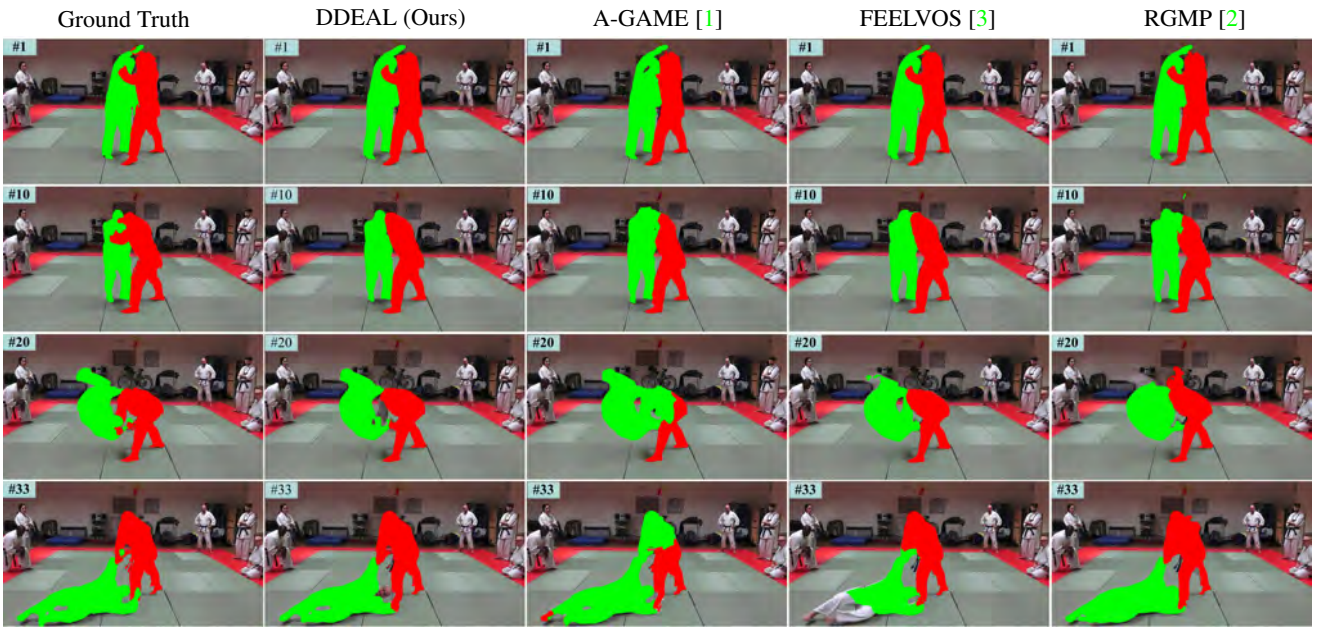

Figure 8: Qualitative results of different methods on the judo sequence of the DAVIS-2017 validation set.

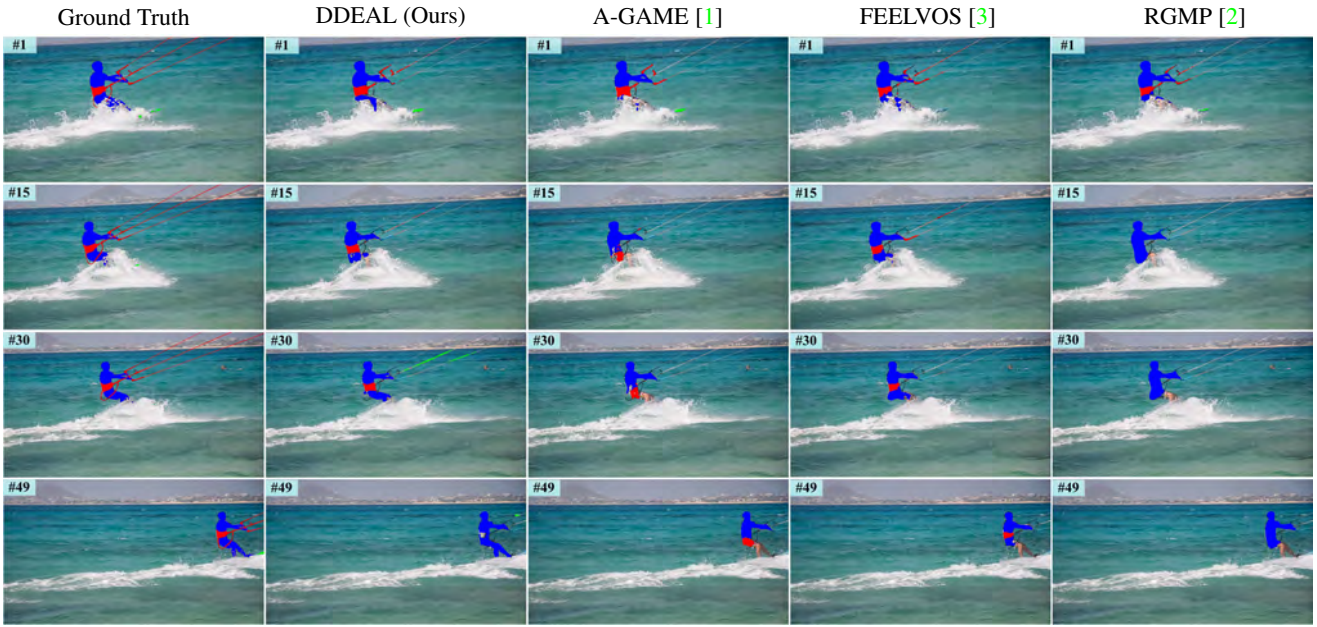

Figure 9: Qualitative results of different methods on the kite-surf sequence of the DAVIS-2017 validation set.

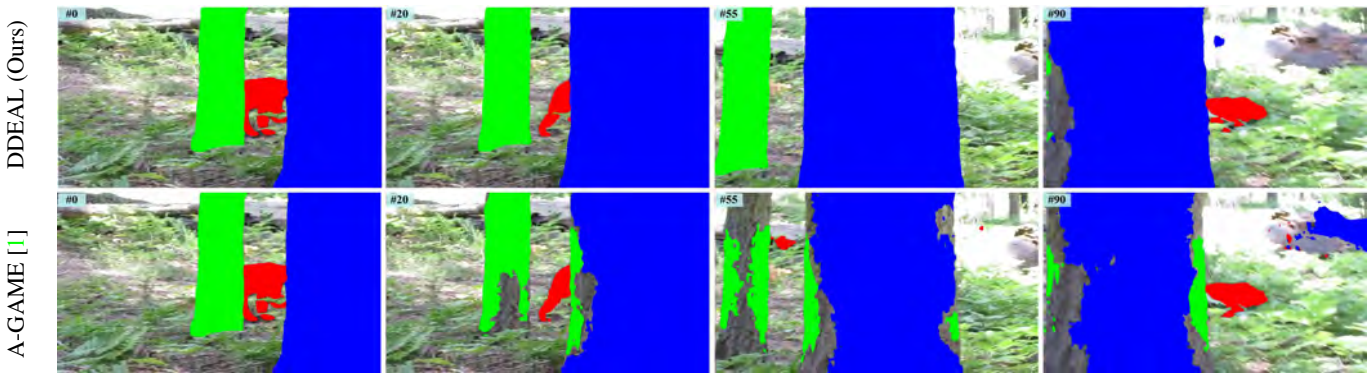

Figure 10: Qualitative results of different methods on the sequence 0a49f5265b of the YouTube-VOS validation set.

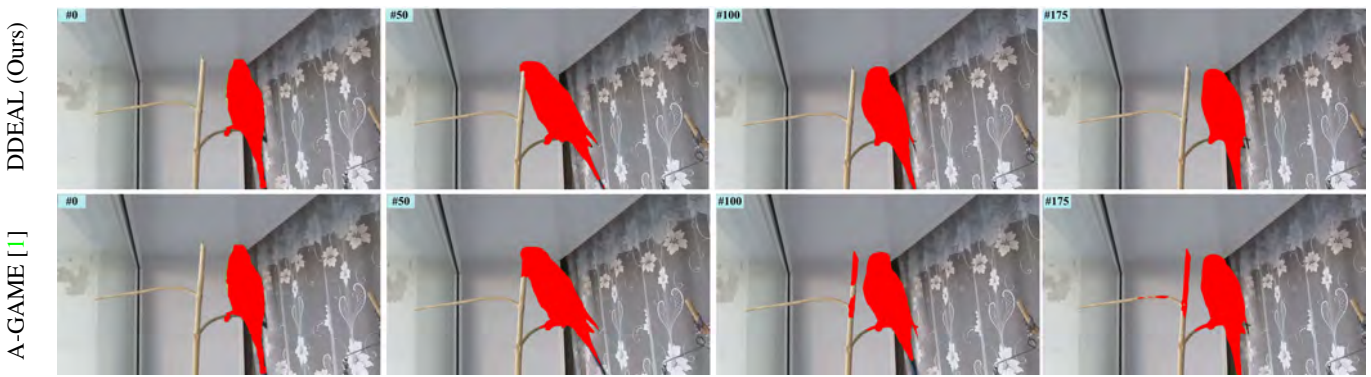

Figure 11: Qualitative results of different methods on the sequence 2ac37d171d of the YouTube-VOS validation set.

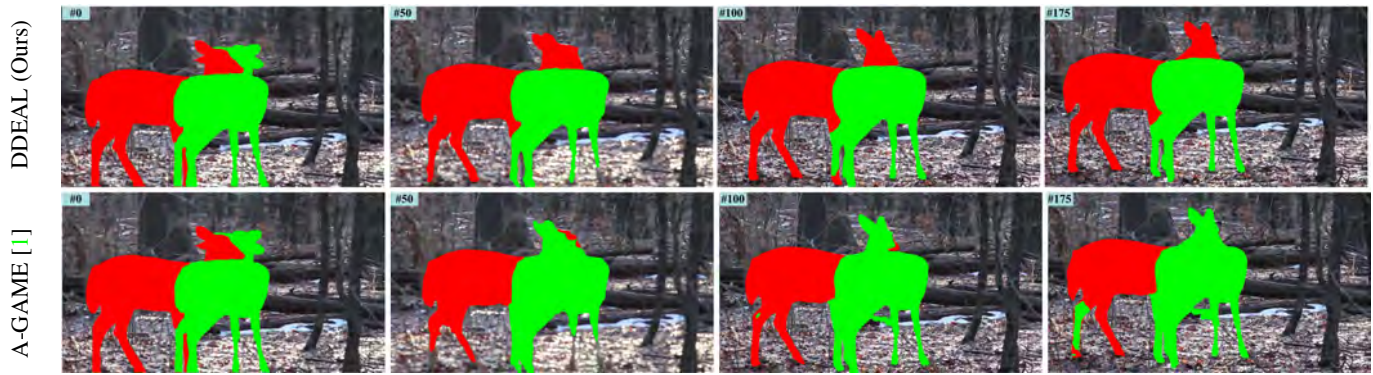

Figure 12: Qualitative results of different methods on the sequence 3f99366076 of the YouTube-VOS validation set.

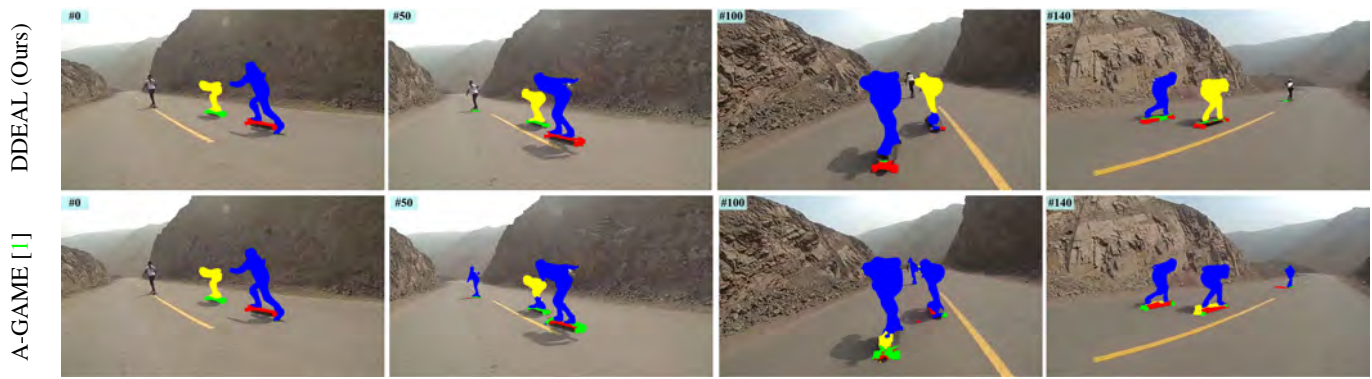

Figure 13: Qualitative results of different methods on the sequence 4bef684040 of the YouTube-VOS validation set.

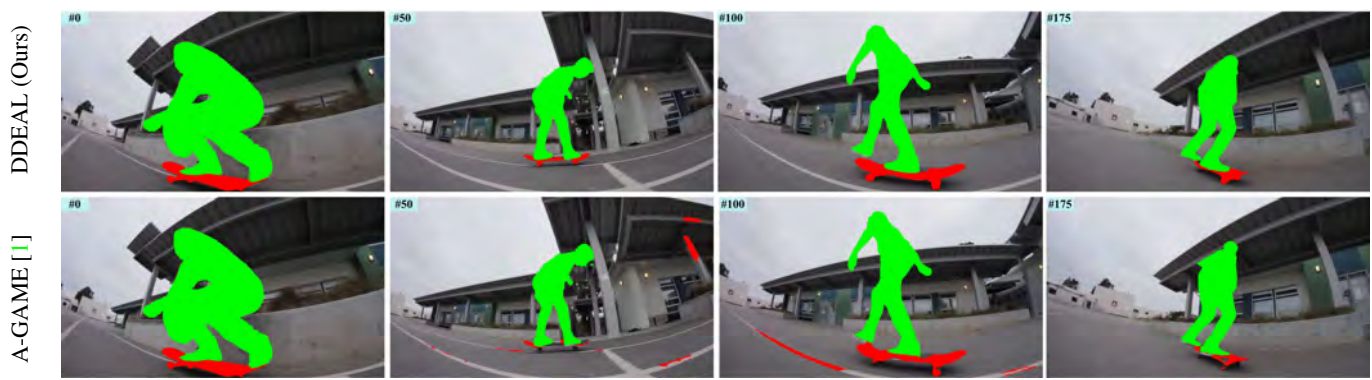

Figure 14: Qualitative results of different methods on the sequence 6ca84fa2b7 of the YouTube-VOS validation set.

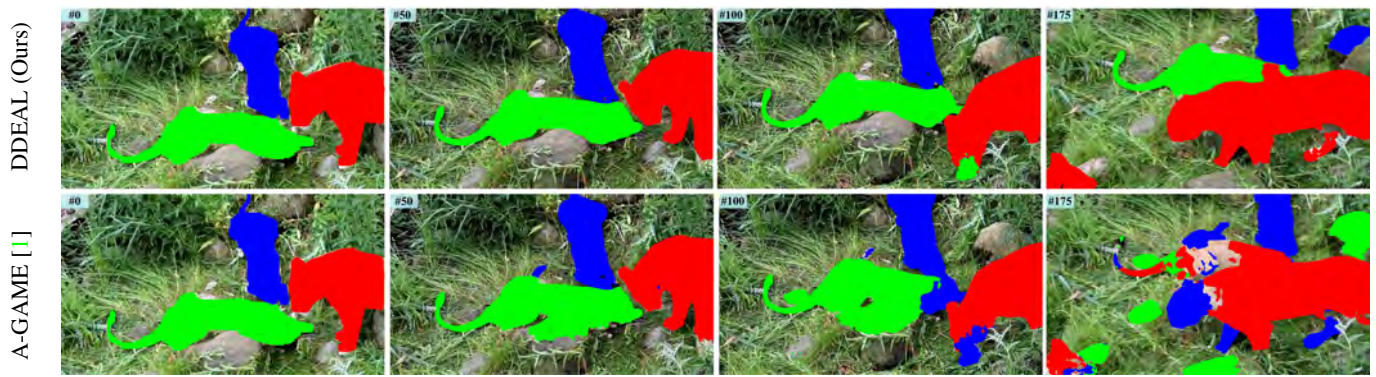

Figure 15: Qualitative results of different methods on the sequence 193aa74f36 of the YouTube-VOS validation set.